# Deep Learning-Based Arrhythmia Detection Using RR-Interval Framed Electrocardiograms

*Song-Kyoo Kim, Chan Yeob Yeun, Paul D. Yoo, Nai-Wei Lo and Ernesto Damiani*

**ABSTRACT**

Deep learning applied to electrocardiogram (ECG) data can be used to achieve personal authentication in biometric security applications, but it has not been widely used to diagnose cardiovascular disorders. We developed a deep learning model for the detection of arrhythmia in which time-sliced ECG data representing the distance between successive R-peaks are used as the input for a convolutional neural network (CNN). The main objective is developing the compact deep learning based detect system which minimally uses the dataset but delivers the confident accuracy rate of the Arrhythmia detection. This compact system can be implemented in wearable devices or real-time monitoring equipment because the feature extraction step is not required for complex ECG waveforms, only the R-peak data is needed. The results of both tests indicated that the Compact Arrhythmia Detection System (CADS) matched the performance of conventional systems for the detection of arrhythmia in two consecutive test runs. All features of the CADS are fully implemented and publicly available in MATLAB.

## I. INTRODUCTION

The use of time-sliced electrocardiogram (ECG) data was originally developed for biometric applications [1-3]. ECGs are usually encountered in a medical setting [4, 5] but are also useful for security because they offer a novel way to identify individuals [6-11]. ECGs monitor the electrical activity of the heart, and can be used to capture analog signal profiles that allow personal identification and authentication when converted into digital data [12]. The components of ECG signals are often used by cardiologists to diagnose cardiovascular problems [13-15]. However, the manual analysis of ECG signals is challenging because it is difficult to categorize the different waveforms and signal morphologies accurately [16]. Automated ECG signal analysis has been introduced to avoid human errors and reduce costs, recently including the use of artificial intelligence [16-18]. Time-sliced ECG data is highly flexible because it can be mixed with other training inputs and is compatible with various ML methods without the need to categorize all the featured waveforms: only the R-peaks are required. The sliced data are used as the input parameters to train neural networks. Two examples are shown in Fig. 1, one based on a fixed time window and the other based on RR-interval framing (RRIF). Unlike conventional time slicing method [1, 2], the proposed RRIF method uses ECG data based on the interval between heartbeats instead of the slicing window time [3]. In this article, we use RRIF as a new approach to provide time-sliced ECG data for ML systems, allowing the detection of arrhythmia. We have designed a Compact Arrhythmia Detection System (CADS) based on a CNN with binary outputs, which requires only short-range ECG data to achieve detection, allowing it to be used with conventional wearable devices or real-time monitoring equipment. The RRIF provides is highly effective for ECG analysis because only the R-peaks are used without detecting other ECG waveform features. It has been applied for an ECG based authentication system by using compact size of data which feed the machine learning engine [3].

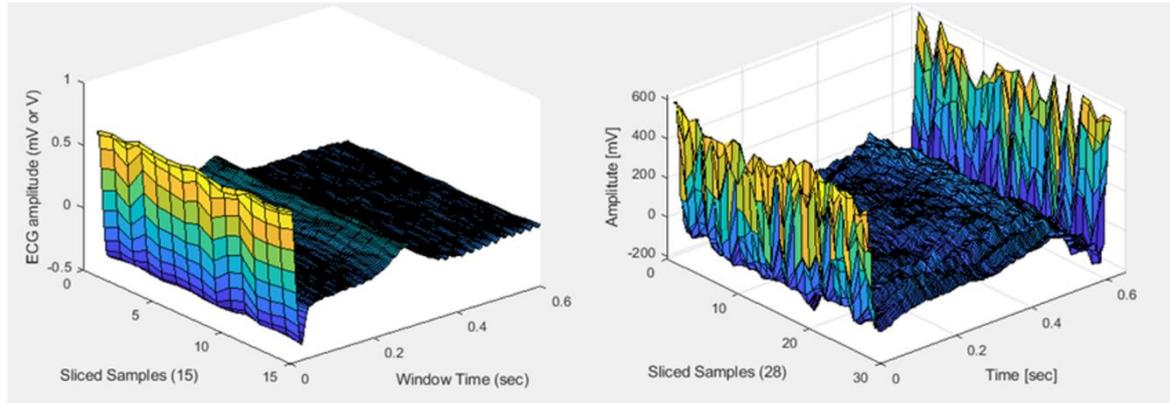

**Figure 1.** Fixed time window (left) vs RR-interval frame (right) for the collection of time-sliced ECG data [35].

## II. COMPACT ARRHYTHMIA DETECTION SYSTEM BASED ON DEEP LEARNING

The analysis and classification of ECG data using various ML techniques, including deep learning, has been studied widely [28-32]. CNNs are often used in the context of deep learning, and CNN variants have proven highly successful in classification tasks across different domains [24-26][33-35]. However, the learning and detection capabilities of CNNs may be insufficient if there is a lot of redundant information, and the analysis of large-scale or highly dimensional data can be computationally demanding. This is realized by alternating convolution and subsampling layers. The last few layers in the CNN use fully-connected MLP-based neural network classifiers to produce the abstracted results. The process flow of the CADS is shown in Fig. 2. Among the standard features of ECG waveforms described by the American National Standards Institute and Association for the Advancement of Medical Instrumentation [38], the CADS uses only the R-peaks.

The process is adapted from previous research by authors and the proposed process for the CADS using CNN is shown in Figure 2. The process starts onto the training phase to acquire corresponding ECG data from both regular and arrhythmia patients as the training dataset. The RRIF method is used on top of these data after filtering ones with higher quality. The RRIF data is used as the input of a CNN engine as a core process mechanism to generate a CADS evaluation model. The core process in this paper supports the trained CNN engine as the evaluation model for a CADS. The NN Engine is generated when the training phase is completed. In the testing phase, the CADS requests associated with newly received ECG data is generated and the RRIF method after filtering is applied to obtain enhanced data. Then the enhanced data are sent to validation process as the input to check with the NN engine for the final decision on this arrhythmia detection request.

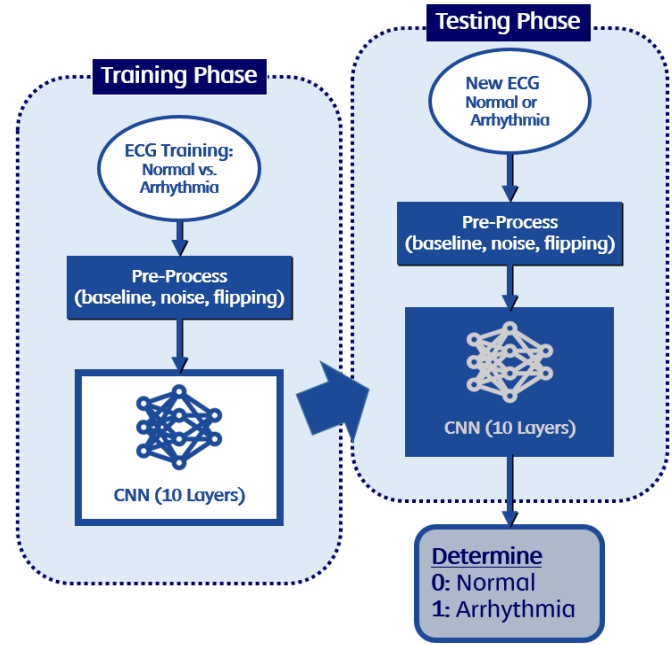

**Figure 2.** The process of the Compact Arrhythmia Detection System (CADS) based on deep learning.

## *2.1. RR-Interval Framed ECG*

The new RRIF method slices ECG data based on R-peaks, making each RR-interval a single frame of the sliced data (Fig. 1). Conventional ECG data slicing uses a fixed-time sliding window, an approach known as segmentation. This is widely used to detect major ECG features, such as the P wave, QRS complex and T wave, allowing the identification of various waveforms for wave feature extraction [18, 38-39]. During segmentation-based ECG analysis [4, 40-43], it is necessary to find the starting point of the P-wave to generate one standard ECG signal [39, 40]. The start of the P-wave is considered as the start of the ECG signal cycle.

The sampling size for each RR-interval can be arbitrary, but it should be consistent when used as input data for a CNN. In the other words, all sliced ECG data should have the same sampling size whereas the cycle time between R-peaks can vary. The sampling values at each position within one RR-interval are used as input parameters for the CNN. In this study, the size of each RRIF was fixed as 220 by referencing a standard ECG signal [19, 39-40]. However, the actual number of input parameters from the sliced ECG becomes 221 when the potential minimum value of 0 is included, and 222 when we also consider the average RR-interval.

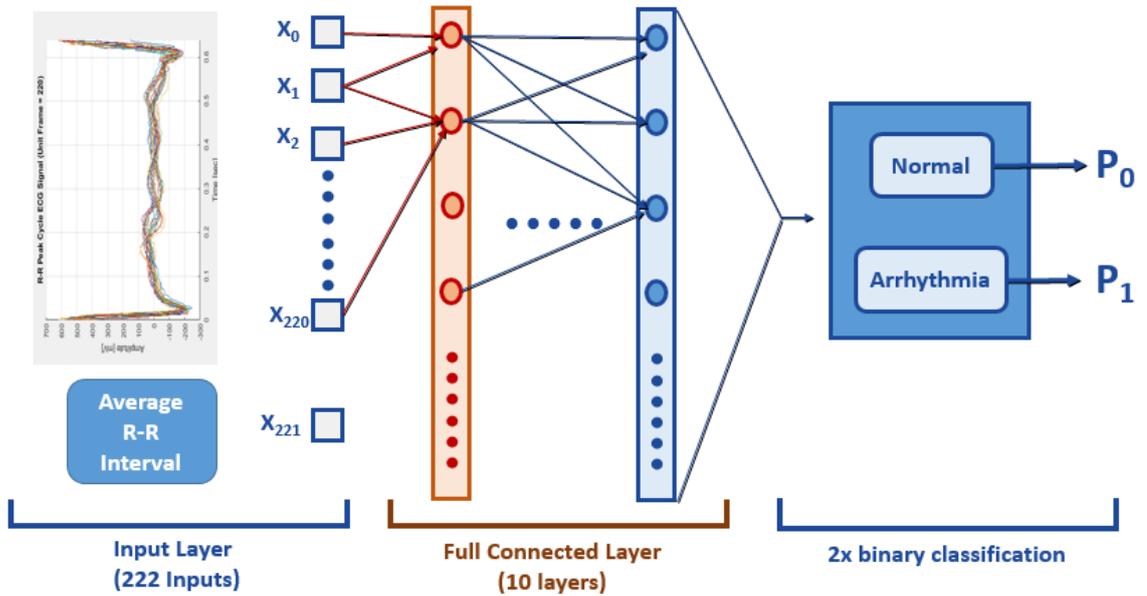

**Figure 3.** Design of a deep learning approach for the Compact Arrhythmia Detection System (CADS).

*2.2. Deep Learning Design for the Detection of Arrhythmia*

Based on the sampling data described above, the CNN architecture used in this study has 222 inputs and a binary output that determines whether the input data represent arrhythmia (positive) or normal cardiac behavior (negative). The structure of the CNN is shown in Fig. 3. This particular architecture does not follow conventional ECG classification standards [38], but focuses solely on the detection of arrhythmia using the RRIF method. The ECG sampling data from the RRIF and the average RR-intervals are used as the input parameters for the CNN. The CNN training dataset was gathered from two sources and comprised 20 normal ECG samples with a high sampling frequency (> 300 Hz) from the Diabetes Complications Research Initiative [45] and 20 arrhythmia ECG samples from the PhysioBank database [46]. Following the CNN training phase, another dataset was used for realistic testing. This involved the same number of samples (20 normal and 20 arrhythmia ECG samples) from different individuals, selected randomly from the ECG testing sample to simulate a realistic situation.

### III. NEURAL NETWORK VALIDATION AND TESTING

The training dataset was validated during CNN training. The training samples comprised 1018 sliced data with 222 input parameters, and the binary output followed the Bernoulli probability distribution. The progress measures of the CNN are shown in Table I. These values were measured automatically using the MATLAB DL function.

**Table I.** Progress of CNN Training for Arrhythmia Detection.

| | |
|---:|:---|
| Epoch | 156 iterations |
| Time | 0.1 s |
| Performance | 0.0594 |
| Gradient | $1.0 \times 10^{-6}$ |
| Validation check | 6 |

The confusion matrix for the training dataset is shown in Table II. The training dataset was divided into three sections for training (70%), internal validation (15%) and testing (15%). This indicates that the training performance (detection of arrhythmia) was around 96%, which is reasonable compared to conventional neural networks.

**Table II.** Confusion Matrix for the CNN Training Phase.

| 1108 sliced pieces (40 persons) | | Actual ECG data | | |
|---|---|---|---|---|
| | | Normal | Arrhythmia | |
| Predicted ECG data | Normal | 528 (51.9%) | 38 (3.7%) | 93.3% 6.7% |
| | Arrhythmia | 6 (0.6%) | 446 (43.8%) | 98.7% 1.3% |
| | | 98.9% 1.1% | 92.1% 7.9% | 95.7% 4.3% |

## IV. CONCLUSIONS

As new ECG sensors become portable for compatibility with smartphones and wearable devices, ECG-based analysis for healthcare applications will become more common. In this article, we have described a novel system (CADS) for the detection of arrhythmia which could be implemented on wearable and portable devices. The main theme of this research is compactness and make the system simple. Hence, our approach involves a new way to build the input parameters for the training of a deep learning algorithm without the need to analyze complex ECG waveforms. Although a CADS is not targeted to handle huge training data for detecting various heart diseases, it is fast and easy to implement into even an IoT device which has limited resources and less computing powers. The RRIF makes this deep learning-based detection system flexible and competitive, achieving 96% accuracy during training and validation and 100% under realistic test conditions.